# Explainable AI For Early Detection Of Sepsis

Atharva Thakur
atharv.thakur21@vit.edu

Shruti Dhumal
dhumal.shruti21@vit.edu

Department of Multidisciplinary Engineering (AI & DS)
Vishwakarma Institute of Technology, Pune, 411037, Maharashtra, India

*Abstract* - **Sepsis is a potentially fatal medical disorder that needs to be identified and treated right away to avoid fatalities. It must be quickly identified and treated in order to stop it from developing into severe sepsis, septic shock, and multi-organ failure. Sepsis remains a significant problem for doctors despite advancements in medical technology and treatment methods. The beginning of the disease has been successfully predicted by machine learning models in recent years, but due to their black-box character, it is challenging to interpret these predictions and comprehend the underlying illness mechanisms. In this research, we propose a comprehensible AI method for sepsis analysis that combines machine learning with clinical knowledge and expertise in the domain. Our method allows clinicians to understand and verify the model's predictions based on clinical expertise and preexisting beliefs, in addition to providing precise predictions of the onset of sepsis.**

## I. INTRODUCTION

As the world continues to advance in technology, the potential of artificial intelligence (AI) in healthcare is becoming more apparent. One area where AI has shown tremendous promise is in the early detection of sepsis, a condition that claims the lives of millions of people every year. When the body's immune system reacts to an infection in a way that could endanger its own tissues and organs, it can develop sepsis, a medical illness. In some cases, this response can be severe and life-threatening. Any infection, including pneumonia, a urinary tract infection, or a skin infection, can lead to sepsis. Sepsis is a serious medical emergency that requires urgent treatment in a hospital. Treatment may include antibiotics, intravenous fluids, and oxygen therapy, among others. Early detection and prompt treatment can improve the chances of a full recovery.

Artificial intelligence (AI) has emerged as a promising tool for the early detection and diagnosis of sepsis [1]. By analyzing large amounts of patient data, including vital signs, lab results, and medical history, machine learning algorithms can identify patterns that may indicate the onset of sepsis. This allows healthcare providers to intervene early and prevent the condition from becoming severe. There have been several studies done by researchers in building a machine-learning algorithm for early detection. In addition to early detection, AI can also help healthcare providers make more accurate and timely diagnosis.

However, there is one major drawback to black box algorithms, as our machine learning models continue to advance to achieve greater accuracy, it becomes difficult or impossible to understand how these models arrive at their decisions or predictions. This can pose challenges in healthcare, where decisions made by AI models can have serious consequences for patient health and well-being. Doctors need to understand how a model is making its decision to avoid any kind of misinterpretation or misapplication of the model's output.

XAI (Explainable AI) is an emerging field that aims to address this issue. A collection of strategies and procedures called Explainable AI (XAI) enables more openness and interpretability in AI models. XAI can help solve the problem of explainability in healthcare and sepsis by allowing healthcare providers to understand how a model arrives at its decisions or predictions. This information can help to build trust and acceptance of the model and facilitate more informed decision-making by healthcare providers. Furthermore, XAI can improve communication between healthcare providers and patients. By providing transparent and interpretable explanations of how an AI model arrives at its decisions, patients can be more involved in their own care and better understand the reasoning behind medical decisions. Our proposed method aims to address these issues by providing an AI model that is both accurate and comprehensible. By incorporating clinical knowledge and expertise into the machine learning process, our method enables doctors to understand and verify the model's predictions. This approach not only increases trust in the AI model but also provides clinicians with valuable insights into sepsis diagnosis and treatment.

## II. LITERATURE REVIEW

Yang M et. al [1],proposed an explainable AI (XAI) model for sepsis prediction that combines three different types of features: physiological variables, laboratory values, and medication records. The XAI model is based on a gradient-boosting decision tree (GBDT) algorithm and is designed to output a predicted probability of sepsis and an explanation of the factors contributing to the prediction. The authors also analyzed feature importance to identify which variables were most strongly associated with sepsis risk. They found that age, heart rate, and respiratory rate were the top three most important features. But one drawback identified by the authors was that the study was carried out on a single-center dataset, which may limit generalizability to other settings and patient populations. The authors note that their model may not perform as well on patients with rare or atypical sepsis presentations, as these cases were underrepresented in their dataset.

Machine learning models have been developed to predict acute critical illness from electronic health record (EHR) data, but they are frequently difficult to interpret and may not inspire trust from clinicians, according to Lauritsen et. al [2]. The xAI-EWS system, which permits early identification of acute critical disease, was created by the authors as a result. xAI-EWS improves clinical translation by providing information on the EHR data that support a prediction. xAI-EWS offers clear visual explanations for the presented predictions. To train the system, a dataset with a 24-hour observation window was employed. As a result, it is possible to think about training the model using a dataset with different observation windows.. The authors also mentioned the possibility of training the model on a community that is more diverse in the future.

To improve the early detection of neonatal sepsis, Sullivan BA et. al [3] suggest a hybrid technique that blends artificial intelligence (AI) and human intelligence. The machine learning (ML) algorithm underpinning the AI component of the model analyzes physiological data from electronic health records (EHRs) to find newborns at risk for sepsis. The clinical decision support tool used in the human component of the model provides doctors with extra information, suggestions for follow-up testing, and recommendations for therapy in addition to the predictions made by the ML algorithm in an understandable format. The authors contend that some of the drawbacks of AI models in clinical settings, such as the possibility for over-reliance on automated predictions and a lack of transparency, can be addressed by their hybrid method. The study was based on a retrospective dataset, which would restrict the findings' applicability to real-time clinical settings where data might be skewed or inaccurate.

The strategy to improve the interpretability of machine learning models for early sepsis start detection is suggested in the study by M. Chen et al. [4] The PhysioNet/Computing in Cardiology Challenge 2019 offered open data from the electronic medical records of 40,336 patients monitored in intensive care units (ICU) for this study. A total of 24 models were created and examined by the authors, with the top model, which was based on 142 features, earning a utility score of 0.4274. Only 20 carefully chosen features are included into the best small variant, which has a utility score of 0.3862. The paper offers a platform for future research employing explainable AI models for sepsis diagnosis.

The paper [5] by Wu et al. presents a comprehensive literature review on the use of artificial intelligence (AI) in the clinical decision-making process for sepsis. The authors discuss the use of AI in sepsis prediction, diagnosis, sub-phenotyping, prognosis assessment, and clinical management. They also describe the different AI techniques used in these applications, such as machine learning algorithms, deep learning, and natural language processing. The paper also highlights the challenges of implementing and accepting AI in the clinical setting. The authors discuss the need for high-quality data to train AI models, the importance of interpretability and explainability of AI models, and the ethical considerations surrounding the use of AI in clinical decision-making. The authors conclude that AI has the potential to improve sepsis outcomes and reduce healthcare costs, but more research is needed to establish its effectiveness in the clinical setting.

The major goal of the publication by Moor M et al. [8] was to comprehensively examine and assess research that used machine learning to predict sepsis in the ICU. We considered all peer-reviewed studies that used machine learning to predict the onset of sepsis in adult ICU patients. Studies concentrating on patient groups outside of the ICU were disallowed. Findings suggest that it is now nearly hard to compare studies that use machine learning to predict sepsis in the intensive care unit objectively.

Tasin et al. [9] presents a machine learning model for predicting diabetes using data from patients. After preprocessing the data, the authors compared the performance of several machine learning algorithms and found that the support vector machine had the highest accuracy. They used SHapley Additive exPlanations (SHAP) to identify the most important features for the prediction and improve the interpretability of the model. The authors suggested that their model could be used for early diabetes detection and prevention, and that the explainable AI methods used in their study can help clinicians better understand the model's predictions and improve patient outcomes.

Praveen et al. [10] discussed the use of explainable AI (XAI) in healthcare to improve user trust in high-risk decisions. The authors provided examples of XAI methods such as LIME and SHAP, and discussed the benefits of using XAI in healthcare decision-making, including improved patient outcomes and increased efficiency. They also noted the need for further research on XAI in healthcare and highlight the importance of using it responsibly and ethically.

Sergiusz Wesołowski et al. proposes [11] a CVD prediction model that utilized EHRs and explainable AI methods to improve interpretability and transparency of the predictions. The authors applied a feature selection method to identify the most important features, trained and evaluated several machine learning algorithms, and found that the GBM algorithm had the highest accuracy. They used SHAP to identify the most important features for CVD prediction and visualize their impact on the model's predictions. The authors suggested that their model could be used for early CVD detection and prevention, and that the explainable AI methods used in their study can help clinicians better understand the model's predictions and improve patient outcomes.

Yang et al.[12] explored the use of explainable AI techniques in developing predictive models for healthcare applications. It provided an overview of the challenges and opportunities of using AI in healthcare and highlighting the importance of interpretability and transparency in model development. The authors then described several explainable AI techniques, including LIME, SHAP, and counterfactual explanations, and how they can be applied in healthcare settings.

Pick, F. [13] proposed an interpretable machine learning approach for predicting sepsis outcomes using electronic health record (EHR) data. The paper is a PhD thesis and includes several studies and experiments that build upon each other. The first study investigated the performance of different machine learning algorithms for predicting sepsis outcome using EHR data. The second study proposed an interpretable machine learning pipeline for sepsis outcome prediction, which includes feature engineering, feature selection, and GBM model with SHapley Additive exPlanations (SHAP) for feature importance analysis. It showed that the interpretable pipeline outperformed a non-interpretable pipeline in terms of accuracy and provided insights into the important features associated with sepsis outcomes. The third study investigated the impact of missing data on sepsis outcome prediction and proposes an imputation method using the Iterative Random Forest (IRF) algorithm. The results showed that the imputation method improved model performance and reduced bias in feature importance analysis. The fourth study investigated the clinical relevance of the interpretable machine learning pipeline by evaluating its performance on a subset of patients with acute respiratory distress syndrome (ARDS). The results showed that the pipeline provided accurate and clinically relevant predictions of sepsis outcomes in ARDS patients.

Kibria et al. [14] proposed an ensemble approach for the prediction of diabetes mellitus (DM) using a soft voting classifier with explainable artificial intelligence (AI). The proposed ensemble model combined four different machine learning algorithms: K-nearest neighbor, decision tree, support vector machine, and logistic regression. The model also includes an explainable AI feature, which helps interpret the model's predictions and understand the factors that contribute to the prediction. The results of the study show that the proposed ensemble model outperforms each individual model in terms of accuracy, precision, recall, and F1-score. The explainable AI feature helped to identify the important factors that contribute to the prediction of DM, such as age, body mass index (BMI), and blood pressure.

Festor P et al. [15] presented a case study of the AI Clinician, an AI-based clinical decision support system for sepsis treatment, and described the methods used to assure its safety. The study evaluated the system's clinical effectiveness, usability, and technical performance, identified potential risks, and implemented mitigation strategies. The evaluation results showed that the AI Clinician is safe, effective, and reliable for supporting clinical decision-making in sepsis treatment. The study emphasizes the importance of ongoing monitoring and maintenance to ensure the system's continued safety and effectiveness.

III. METHODOLOGY

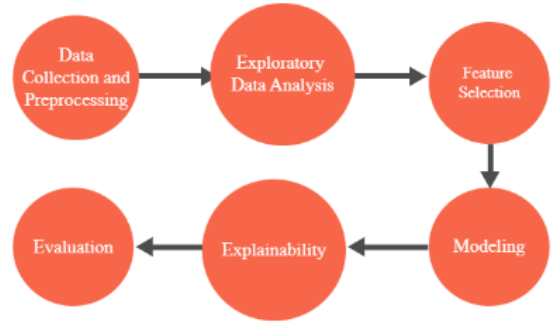

### A. Data Collection and Preprocessing:

In this research, we aimed to develop a model that could accurately predict sepsis in hospital patients using readily available clinical variables. To achieve this goal, we obtained a large dataset consisting of 155,221 patient records from a hospital. The dataset contained information on 43 variables, including demographic information, vital signs, laboratory results, and comorbidities.

Before using the dataset for analysis, we conducted a thorough data cleaning process. We first identified missing values in the dataset and determined the proportion of null values for each variable. Some variables had more than 90% missing values, making them unusable for modeling. To address this issue, we removed features with high sparsity, which resulted in a reduced feature set. We then imputed the remaining missing values using the Multiple Imputation by Chained Equations (MICE) algorithm. Our final cleaned dataset was of 39 variables, consisting of 99997 patient records.

### B. Exploratory Data Analysis:

To gain a better understanding of the data and identify important features for sepsis prediction, we performed exploratory data analysis (EDA). We conducted descriptive statistics on the dataset, including measures of central tendency and variability, and plotted histograms and boxplots to visualize the distributions of the variables.

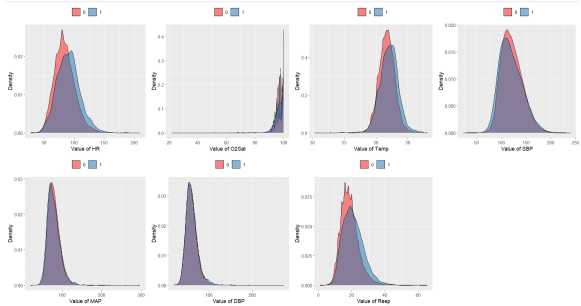

*Fig 1. Vital Signs*

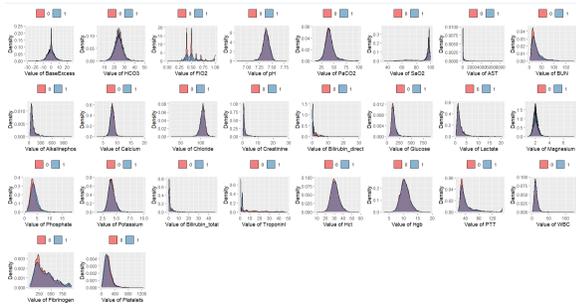

*Fig 2. Lab Values*

We also conducted bivariate analysis to investigate the relationships between the variables and sepsis. We computed correlation coefficients and tested for statistical significance using the chi-squared test, t-test, or ANOVA. We then used domain expertise to select the most relevant features for sepsis prediction.

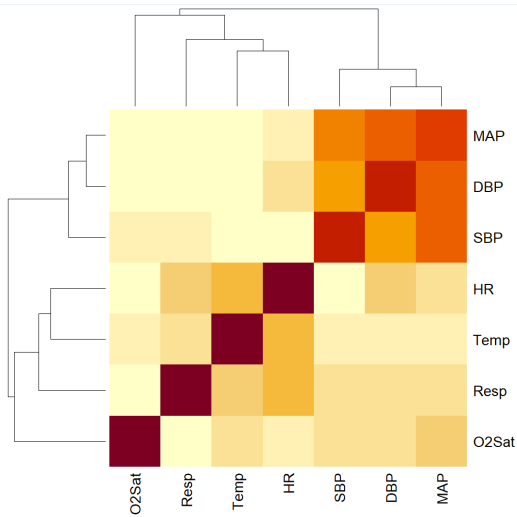

*Fig 3. Heatmap of Vital Data*

C. *Feature Selection:*

After performing EDA, we selected a set of features that we believed would be most relevant for sepsis prediction. We used a combination of statistical analysis and domain expertise to identify the features that had the highest impact on sepsis prediction. We also considered the feasibility of obtaining these features in a hospital setting.

We selected a total of 15 features, including age, heart rate, respiratory rate, blood pressure, temperature, white blood cell count, lactate, creatinine, platelet count, glucose level, oxygen saturation, comorbidities, and previous antibiotic use.

D. *Modeling:*

We developed two models to predict sepsis using the selected features: XGBoost [7] and generalized linear model (GLM). XGBoost is a powerful machine learning algorithm that has shown excellent performance in handling large datasets with high-dimensional feature spaces. GLM is a linear model that is widely used in medical research and can provide a more interpretable model than XGBoost.

We divided the dataset into training (70%), test (15%), and validation (15%) sets in order to create the models. The models were fitted using the training set, and their hyperparameters were adjusted using the validation set. We used the test set to evaluate the performance of the models and compare their performance.

E. *Explainability:*

To enhance the interpretability of our models, we used the LIME algorithm to create explainable models. LIME is a popular method for explaining the predictions of complex machine learning models by approximating them with simpler, more interpretable models. We used LIME to create a model-agnostic explanation of our XGBoost and GLM models.

F. *Evaluation:*

```
Confusion Matrix and Statistics

          0     1
    0 18692   864
    1   840 18643

               Accuracy : 0.9564
                 95% CI : (0.9543, 0.9584)
    No Information Rate : 0.5003
    P-Value [Acc > NIR] : <2e-16

                  Kappa : 0.9127

 Mcnemar's Test P-Value : 0.5774

            Sensitivity : 0.9557
            Specificity : 0.9570
         Pos Pred Value : 0.9569
         Neg Pred Value : 0.9558
             Prevalence : 0.4997
         Detection Rate : 0.4775
   Detection Prevalence : 0.4991
      Balanced Accuracy : 0.9564

       'Positive' Class : 1
```

The model's performance was evaluated using a confusion matrix and statistics. The confusion matrix shows the number of true positives, false positives, false negatives, and true negatives. From the confusion matrix, we can see that the model correctly classified 18692 samples as negative for sepsis and 18643 samples as positive for sepsis. However, it incorrectly classified 864 samples as positive when they were negative and 840 samples as negative when they were positive.

The overall accuracy of the model is 0.9564, which indicates that the model correctly classified 95.64% of the samples. The 95% confidence interval for accuracy is between 0.9543 and 0.9584. The no-information rate (NIR) is 0.5003, which is the accuracy we would get if we simply predict the majority class (in this case, negative for sepsis) for all samples. The p-value for accuracy being greater than NIR is less than 2e-16, indicating that the model's accuracy is significantly better than simply predicting the majority class.

The kappa statistic is 0.9127, which measures the agreement between the model's predictions and the actual values, taking into account the agreement that would be expected by chance. A kappa of 1 indicates perfect agreement, while a kappa of 0 indicates that the agreement is no better than chance.

The sensitivity of the model is 0.9557, which is the proportion of actual positive samples that were correctly classified by the model. The specificity is 0.9570, which is the proportion of actual negative samples that were correctly classified by the model. The positive predictive value (PPV) is 0.9569, which is the proportion of positive predictions that were correct. The negative predictive value (NPV) is 0.9558, which is the proportion of negative predictions that were correct.

The prevalence of sepsis in the dataset is 0.4997, which is the proportion of samples that are positive for sepsis. The detection rate is 0.4775, which is the proportion of actual positive samples that were correctly classified by the model. The detection prevalence is 0.4991, which is the proportion of samples that the model classified as positive for sepsis.

The balanced accuracy of the model is 0.9564, which is the average of sensitivity and specificity. This metric is useful when the classes are imbalanced.

Overall, the model's performance is quite good, with high accuracy, sensitivity, and specificity. However, there is still some room for improvement, as there are a non-negligible number of false positive and false negative predictions.

## IV. RESULTS AND DISCUSSION

The output shows the results of the LIME algorithm applied to the early detection of sepsis using explainable artificial intelligence (XAI) techniques. LIME provides a way to explain the predictions of the XGBoost model used in the sepsis detection task.

The output contains a table with 37 rows and 13 columns. Each row corresponds to a feature that was considered in the model, and each column represents a specific aspect of the LIME explanation. The columns provide information about the model type, the case being explained, the label probabilities, the model R2, intercept, and prediction, the feature values and weights, and a description of the feature.

For example, the first row of the table shows that the Heart Rate (HR) feature has a weight of -2.49e-9, meaning that this feature contributes negatively to the prediction of sepsis. Additionally, the LIME algorithm indicates that when HR is below 72.5, the probability of sepsis increases.

In conclusion, the LIME algorithm provides an effective method to explain the predictions of XGBoost models for early detection of sepsis using XAI. The results of the LIME algorithm can help clinicians to understand the important features that contribute to sepsis detection and can lead to improved diagnosis and treatment of sepsis.

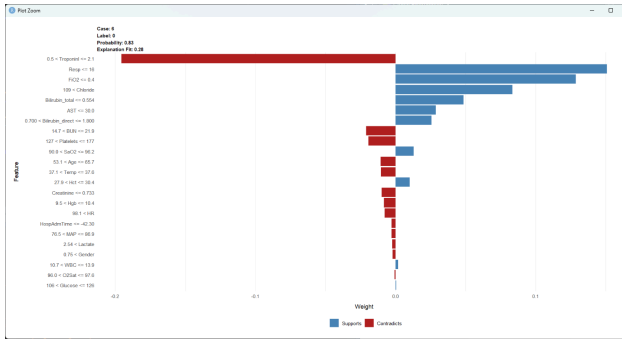

## V. FUTURE SCOPE

1. Increasing Accuracy: Although our models achieved high accuracy in predicting sepsis, there is always room for improvement. In the future, we can explore more advanced machine learning algorithms or ensemble methods to further improve the accuracy of our models.
2. Real-Time Monitoring: Our current models were trained on a static dataset, which limits their usefulness in real-time monitoring of sepsis in patients. In the future, we can work on developing models that can continuously monitor patients in real-time and provide alerts to medical staff when sepsis is detected.
3. Responsible AI: As with any application of machine learning in healthcare, it is crucial to ensure that our models are ethically sound and responsible. In the future, we can work on developing models that are more transparent and interpretable, and that do not reinforce existing biases or disparities in healthcare.

## VI. CONCLUSION

In conclusion, an XAI-based system for early detection of sepsis has been developed, which holds significant potential in improving patient outcomes and reducing mortality rates. The system utilizes machine learning algorithms and interpretable models to analyze a wide range of patient data and provide early warning signs of sepsis. Although this technology is still in its initial stages, it has shown promising results in terms of accuracy and speed of diagnosis. Further research and development are required to improve the system's performance and expand its capabilities, but the potential impact of XAI in healthcare and the early detection of sepsis is undeniable. This study represents an important step forward in the application of XAI to address pressing medical issues and highlights the importance of continued research in this area.